\documentclass{article}
\PassOptionsToPackage{numbers,sort&compress}{natbib} % Ensure natbib orders citations as they appear

\usepackage[preprint]{neurips_2024}
\usepackage[utf8]{inputenc} % allow utf-8 input
\usepackage[T1]{fontenc}    % use 8-bit T1 fonts
\usepackage{hyperref}       % hyperlinks
\usepackage{url}            % simple URL typesetting
\usepackage{booktabs}       % professional-quality tables
\usepackage{amsfonts}       % blackboard math symbols
\usepackage{nicefrac}       % compact symbols for 1/2, etc.
\usepackage{microtype}      % microtypography
\usepackage{xcolor}         % colors
\usepackage[utf8]{inputenc}
\usepackage{amsmath}
\usepackage{tikz}

\usepackage{hyperref}
\hypersetup{
    colorlinks=true,
    linkcolor=perfred,     % Choose your preferred link color
    citecolor=perfblue,     % Choose your preferred citation link color
    filecolor=magenta,  % Choose your preferred file color
    urlcolor=perfred,      % Choose your preferred URL color
    pdfborder={0 0 0}   % Removes the border around the links
}

\usepackage{url}
\usetikzlibrary{arrows.meta,bending,positioning}
\usepackage[utf8]{inputenc} % allow utf-8 input
\usepackage[T1]{fontenc}    % use 8-bit T1 fonts
\usepackage{url}            % simple URL typesetting
\usepackage{booktabs}       % professional-quality tables
\usepackage{amsfonts}       % blackboard math symbols
\usepackage{nicefrac}       % compact symbols for 1/2, etc.
\usepackage{microtype}      % microtypography
\usepackage{microtype}
\usepackage{graphicx}
\usepackage{subfigure}
\usepackage{tabularx}
\usepackage{wrapfig}
\usepackage{array}
\usepackage{microtype}
\usepackage{colortbl}      % To allow color in tables

\usepackage{caption}
\usepackage{amsfonts}       % blackboard math symbols
\usepackage{amsmath}
\usepackage{amssymb}
\usepackage{mathtools}
\usepackage{amsthm}
\usepackage{bbm}
\usepackage{lipsum}

\usepackage{algorithm}
\usepackage{algpseudocode}
\usepackage{adjustbox}
\usepackage[font=small]{caption}

\usepackage{fancyvrb}
\usepackage{float}         % for floating figures and algorithms
\usepackage{enumitem}
\usepackage{multirow}
\usepackage{rotating}
\usepackage{xspace}
\usepackage{soul}
\usepackage[toc,page,header]{appendix}
\usepackage{minitoc}

\usepackage{listings}
\usepackage{float}

\definecolor{codegray}{rgb}{0.5,0.5,0.5}
\definecolor{codepurple}{rgb}{0.58,0,0.82}
\definecolor{backcolour}{rgb}{0.95,0.95,0.92}
\lstdefinestyle{mystyle}{
    backgroundcolor=\color{white},       % Set background color to white
    commentstyle=\color{codegray}\itshape, % Italic comments
    keywordstyle=\color{blue}\bfseries,  % Bold keywords in blue
    numberstyle=\tiny\color{gray},       % Line numbers in tiny font and gray color
    stringstyle=\color{codepurple},      % Strings in purple
    basicstyle=\ttfamily\footnotesize,   % Monospaced font, footnote size
    breakatwhitespace=false,             % Do not break lines at whitespace
    breaklines=true,                     % Automatically break lines
    captionpos=b,                        % Position caption at the bottom
    keepspaces=true,                     % Keep spaces in code
    numbersep=5pt,                       % Space between numbers and code
    showspaces=false,                    % Don't show spaces explicitly
    showstringspaces=false,              % Don't show spaces in strings
    showtabs=false,                      % Don't show tab characters
    tabsize=2,                           % Set tab size to 2 spaces
    frame=shadowbox,                         % Adds a border (alternatives: 'single', 'shadowbox')
    rulecolor=\color{gray},              % Border color set to gray
    framesep=2mm,                        % Space between code and border
    xleftmargin=2mm,                     % Left margin for indentation
    xrightmargin=2mm,                    % Right margin for indentation
    aboveskip=2mm,                       % Space above code block
    belowskip=2mm,                       % Space below code block
    frameround=tttt                      % Optional: rounded corners (can be removed for straight corners)
}
\usepackage[para]{footmisc}

\graphicspath{{figures/}}

\algnewcommand{\algcomment}[1]{\textit{// #1}}  % custom 
\newcommand{\figGap}[0]{\vspace{-\baselineskip}}

\definecolor{lightgray}{gray}{0.9}
\definecolor{lightgreen}{rgb}{0.8,1,0.8}
\definecolor{perfblue}{RGB}{64, 114, 175}
\definecolor{perfred}{RGB}{220, 60, 60}

\newcommand{\cellgreen}{\cellcolor{lightgreen}}

\newcommand{\react}{\texttt{ReAct}\xspace}
\newcommand{\gptfo}{\texttt{gpt-4o}\xspace}
\newcommand{\gptfomini}{\texttt{gpt-4o-mini}\xspace}

\theoremstyle{plain}

\theoremstyle{definition}

\theoremstyle{remark}

\newcommand{\agentprm}{\texttt{AgentPRM}\xspace}
\newcommand{\agentprms}{\texttt{AgentPRMs}\xspace}
\newcommand{\inverseprm}{\texttt{InversePRM}\xspace}

\title{Process Reward Models for LLM Agents: \\ 
Practical Framework and Directions}

\author{%
  Sanjiban Choudhury \\
  Cornell University \\
  \texttt{sanjibanc@cornell.edu} \\
}

\begin{document}

\maketitle

\begin{abstract}
We introduce Agent Process Reward Models (\agentprm), a simple and scalable framework for training LLM agents to continually improve through interactions. \agentprm follows a lightweight actor-critic paradigm, using Monte Carlo rollouts to compute reward targets and optimize policies. It requires minimal modifications to existing RLHF pipelines, making it easy to integrate at scale. Beyond \agentprm, we propose \inverseprm, which learns process rewards directly from demonstrations without explicit outcome supervision. We also explore key challenges and opportunities, including exploration, process reward shaping, and model-predictive reasoning. We evaluate on ALFWorld benchmark, show that small 3B models trained with \agentprm and \inverseprm outperform strong GPT-4o baselines, and analyze test-time scaling, reward hacking, and more. Our code is available at: \url{https://github.com/sanjibanc/agent_prm}. 
\end{abstract}

\section{Introduction}

Large language model (LLM) agents excel in decision-making tasks such as web navigation~\citep{sodhi2024step}, robotics~\citep{black2024pi_0}, and interactive code generation~\citep{jimenez2023swe}. However, they rely heavily on prompting~\citep{yao2022react, shinn2023reflexion} or supervised fine-tuning (SFT~\citep{chen2023fireact}. Prompting demands extensive manual effort~\citep{khot2022decomposed, sodhi2024step} and does not enable autonomous improvement. SFT, while effective, is constrained by demonstration quality and lacks mechanisms for self-correction at test time.

This raises a fundamental question: \emph{How can LLM agents improve through interaction without extensive human supervision?} Reinforcement learning (RL) naturally enables policy refinement through experience, but applying RL to LLM agents presents key challenges: (1) \emph{Long-horizon decision-making:} LLM agents must reason over multiple steps, producing structured multi-token outputs that blend reasoning and actions. (2) \emph{Sparse rewards:} Feedback is often delayed until the end of long interactions, complicating credit assignment. While large-scale RL approaches have been explored~\cite{guo2025deepseek}, they remain impractical due to high sample complexity.

\begin{figure*}[!t]
\centering
\includegraphics[width=\linewidth]{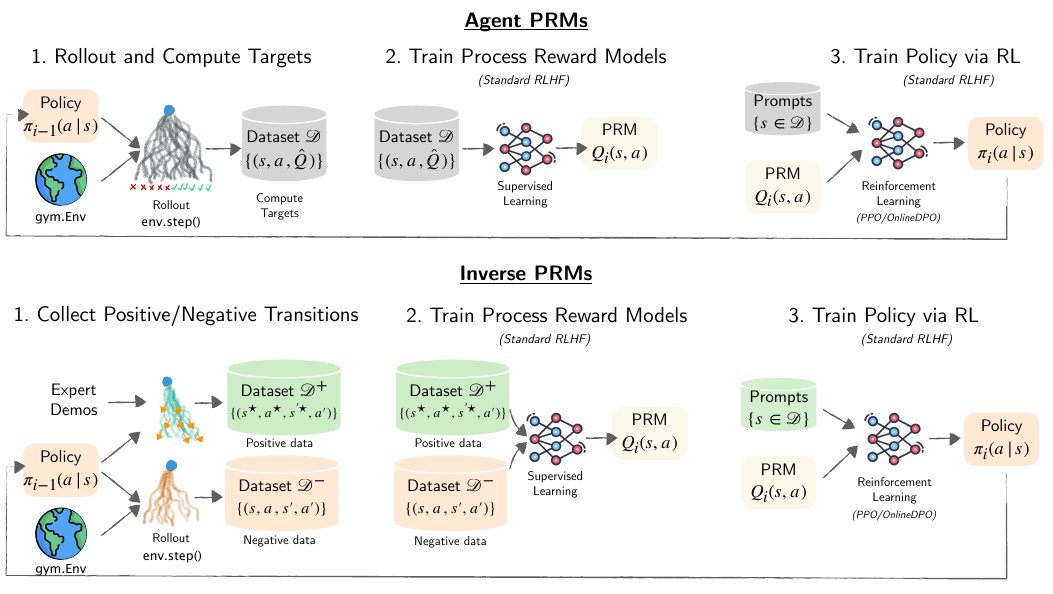}
\caption{\small
\textbf{Overview}  (a) \agentprm: Trains an LLM policy $\pi$ using outcome rewards through three iterative stages. Stage 1: Roll out the current policy $\pi_{i-1}$ and compute the PRM target dataset $\mathcal{D}$. Stage 2: Train PRM $Q_i$ on $\mathcal{D}$ via supervised learning. Stage 3: Update policy $\pi_i$ using RL with PRM $Q_i$.
(b) \inverseprm: Trains $\pi$ using expert demonstrations in three stages. Stage 1: Roll out $\pi_{i-1}$ to generate positive $\mathcal{D}^+$ and negative $\mathcal{D}^-$ transition datasets. Stage 2: Train PRM $Q_i$ to distinguish between $\mathcal{D}^+$ and $\mathcal{D}^-$. Stage 3: Optimize $\pi_i$ via RL with PRM $Q_i$.
Note: Stages 2 and 3 align with standard RLHF pipelines; only Stage 1 is newly introduced. \figGap
}
\label{fig:framework}
\end{figure*}

Instead of large-scale RL, we propose a more tractable alternative: \emph{Agent Process Reward Models} (\agentprm). PRMs provide fine-grained supervision at each step, akin to critic~\cite{haarnoja2018soft} or value functions in RL. By evaluating intermediate actions rather than relying on sparse outcome rewards, PRMs improve sample efficiency. While PRMs have been explored in multi-step reasoning tasks~\citep{lightman2023let, uesato2022solving, setlur2024rewarding}, they are underexplored in agentic settings where actions impact an external environment. Our work addresses this gap.

We propose a simple and scalable framework for training \agentprms. It has two key aspects:
\begin{enumerate}[leftmargin=0.2in, nosep]
\item \emph{Automatic PRM annotation:} PRM targets are computed using asynchronous Monte Carlo rollouts, enabling agents to learn without manually labeled rewards.
\item \emph{Iterative training:} PRMs and policies are jointly trained in an iterative process, where each refines the other to improve overall performance.
\end{enumerate}
The framework is \emph{simple}: it follows the actor-critic paradigm, a well-established RL algorithm with strong theoretical foundations and practical flexibility. 
The framework is \emph{scalable}: it seamlessly integrates into existing RLHF infrastructure~\cite{lambert2024tulu3, vonwerra2022trl} with only one additional component—automatic reward annotation. 

This simple framework opens up new questions, algorithms, and research directions. We introduce \inverseprm, which learns PRMs directly from demonstrations without explicit outcome rewards. \inverseprm achieves higher sample efficiency than \agentprm without added complexity. We also examine challenges in scaling \agentprm, including exploration, sample efficiency, and model-predictive reasoning. To address these, we explore a combination of established RL techniques—such as reset distribution and reward shaping—with LLM-driven strategies like steered exploration and model-predictive reasoning.

Our key contributions are:
\begin{enumerate}[leftmargin=0.2in]
    \item \textbf{Algorithms and Code.} We introduce \agentprm (Sec.~\ref{sec:agent_prm}),  a scalable method for training process reward models, and \inverseprm (Sec.~\ref{sec:inverse_prm}), which learns PRMs directly from demonstrations. Our implementation is a light-weight Gym wrapper around OpenInstruct\footnote{\url{https://github.com/allenai/open-instruct}~\citep{lambert2024tulu3}}, making it easy to integrate with existing RLHF pipelines.  

    \item \textbf{Evaluation and Analysis.} We evaluate on a text game benchmark ALFWorld~\cite{shridhar2020alfred} and find:
    \begin{itemize}
        \item \agentprm enables small (3B) models to outperform strong GPT-4o baselines. We analyze training curves, test-time scaling, reward hacking, and absolute v/s relative loss (Sec.~\ref{sec:agent_prm:exp}).
        \item \inverseprm achieves near expert performance in a single iteration, significantly outperforming SFT and being more sample-efficient than \agentprm (Sec.~\ref{sec:inverse_prm:exp}).
    \end{itemize}

    \item \textbf{Challenges and Opportunities.} We discuss challenges and new research opportunities in:
    \begin{itemize}
        \item \emph{Exploration}: We explore resets and steered exploration to accelerate training (Sec.~\ref{sec:challenge:exploration}).
        \item \emph{Process Reward Shaping}: We use reference policies to shape process rewards to stabilize training in low sample regimes   (Sec.~\ref{sec:challenge:prm}).
        \item \emph{Model-Predictive Reasoning}: We discuss reasoning as model-predictive planning to practically enable large-scale RL to apply to agent settings (Sec.~\ref{sec:challenge:model}).
    \end{itemize}
\end{enumerate}

\section{Agent Process Reward Models: A Simple Framework}
\label{sec:agent_prm}
\subsection{Formulation}
Consider an agent interacting with an environment over multiple turns to solve a task. We model this interaction as a \emph{turn-level} Markov Decision Process (MDP). At turn $t$, the state $s_t$ is the history of observations and actions, $s_t = \{o_0, a_0, \dots, o_{t-1}\}$. The agent selects an action $a_t$ and transitions to a new state $s_{t+1}$ according to the environment dynamics. The agent receives a reward $r(s_t, a_t) \in [0, 1]$, typically provided at terminal states and referred to as the \emph{outcome reward}, which evaluates the overall success of the task.
The agent’s behavior is determined by a policy $\pi(a_t \mid s_t)$, which maps states to a distribution over actions. The objective of the policy is to maximize the expected return, defined as the sum of discounted rewards $ \mathbb{E}_{\pi} \left[ \sum_{t=0}^{T-1} \gamma^t r(s_t, a_t) \right]$, where $\gamma$ is the discount factor.

For LLM agents, each action $a_t$ consists of a sequence of tokens, encoding both reasoning and an environment action. This induces a two-level decision hierarchy: 
\begin{enumerate}[leftmargin=0.2in, nosep] 
\item \emph{Turn-level MDP}: Models the sequence of agent-environment interactions over multiple turns. 
\item \emph{Token-level MDP}: Models the sequence of tokens within each turn, each token is an action. 
\end{enumerate}
Typically, RLHF frameworks are single-turn and hence perform RL only at token-level MDP. We next look at how to lift these frameworks to solve turn-level MDPs. 

\paragraph{Agent Process Reward Models.} 
A \emph{process reward model (PRM)}~\cite{lightman2023let} assigns turn-wise scores in a multi-turn response, providing structured feedback to guide policy learning. In turn-level MDPs, a PRM functions as a state-action value function, analogous to a Q-function in RL. Formally, the PRM is $Q^\pi(s_t, a_t) = \mathbb{E}_\pi\left[\sum_{k=t}^T \gamma^{k-t} r(s_k, a_k) \mid s_t, a_t\right]$. Maximizing PRM $Q^\pi(s_t, a_t)$ enables the policy to improve task performance through intermediate feedback rather than relying on outcome rewards.

\paragraph{Distinction from Reasoning Tasks.} 
PRMs have primarily been studied in multi-step math reasoning tasks~\citep{lightman2023let, wang2024math} where transitions are deterministic and known. In these settings, test-time search methods like beam search~\citep{snell2024scaling} can be used to optimize reasoning sequences. In contrast, LLM agents operate in external environments with unknown, stochastic transitions, where actions have uncertain effects. This makes beam search impractical, as future states cannot be enumerated in advance. We focus on training PRMs and policies under these complex settings.

\subsection{Approach}
We adopt a policy iteration framework to jointly train the process reward model $Q^\pi(s,a)$ and the agent policy $\pi(a|s)$. Algorithm~\ref{alg:prm_training} describes the three-stage process:
\begin{enumerate}[leftmargin=0.2in, nosep]
\item Rolling out the current policy $\pi_\theta$ to collect data and compute Q targets
\item Train the PRM $Q_\phi(s,a)$ given Q targets (\emph{standard RLHF})
\item Train the policy $\pi_\theta$ using reinforcement learning on the trained RM (\emph{standard RLHF})
\end{enumerate} 
This follows standard RLHF pipelines, with the key difference being \emph{Stage 1}, where PRM targets are computed from rollouts rather than preference labels. We describe each stage below.

\paragraph{Stage 1: Rollout and Compute Target.} At iteration $i$, we roll out the policy $\pi_{i-1}$ in the environment to generate trajectories of states, actions, and rewards $\mathcal{D}_{\rm rollout} = \{(s_0,a_0,r_0, \dots, s_{T-1}, a_{T-1}, r_{T-1})\}$. To scale up data collection, we run environments in parallel and step through them in batched mode. Each batch of states is sent to the model, which returns a corresponding batch of actions. We leverage fast inference libraries such as SG-Lang~\cite{zheng2024sglang} and VLLM~\cite{kwon2023efficient}. To improve state coverage, we roll out $\pi_{i-1}$ multiple times on the same task, ensuring repeated state visits. Rollouts are stored in a dictionary $\mathcal{G}(s,a)$, which maps each hashed state-action pair to the set of trajectories passing through $(s,a)$. We compute PRM targets as
\begin{equation}
    \hat{Q}(s, a) = \frac{1}{|\mathcal{G}(s, a)|} \sum_{(s_t, a_t) \in \mathcal{D}(s, a)} \sum_{k=t}^{T-1} \gamma^{k-t} r_k
\end{equation}
Finally we normalize the targets $\hat{Q}(s, a)$ to be between $[0,1]$. The final dataset is then $\mathcal{D} = \{(s, a,\hat{Q})\}$ which is used to train the PRM. Note that we found this approach to be significantly simpler than doing a Monte-Carlo Tree Search (MCTS) which requires synchronous exploration and is difficult to scale. In contrast, we collect our rollouts asynchronously.

\begin{algorithm}[!t]
\caption{Training Agent PRMs}
\label{alg:prm_training}
\begin{algorithmic}[1]
\State Initialize with agent policy $\pi_0$
\For{iteration $i = 1, \dots, K$}
    \Statex
    \Comment{\textbf{Stage 1: Rollout and Compute Targets}}
    \State Collect rollout $\{(\dots, s_t,a_t,r_t,\dots)\}$ using $\pi_{i-1}$ and store in dictionary $\mathcal{G}(s,a)$
    \State Compute PRM targets $\hat{Q}(s, a) = \frac{1}{|\mathcal{G}(s, a)|} \sum_{(s_t, a_t) \in \mathcal{D}(s, a)} \sum_{k=t}^{T-1} \gamma^{k-t} r_k$
    \State Aggregate data into dataset $\mathcal{D} = \{(s, a, \hat{Q})\}$

    \Comment{\textbf{Stage 2: Train Process Reward Model}}
    \State Train PRM $Q_i = \arg\min \mathcal{L}(Q_\phi)$ by minimizing the soft binary cross-entropy loss:
    \begin{equation}
        \mathcal{L}(Q_\phi) = -\mathbb{E}_{(s, a, \hat{Q}) \sim \mathcal{D}} \left[ \hat{Q} \log Q_\phi(s, a) + (1 - \hat{Q}) \log (1 - Q_\phi(s, a)) \right].
    \end{equation}

    \Comment{\textbf{Stage 3: Train Policy via RL}}
    \State Update policy $\pi_i$ to maximize $Q_i$ while regularizing to $\pi_{i-1}$:
    \begin{equation}
    \pi_{i} = \arg\max_{\pi_\theta} \mathbb{E}_{s \sim \mathcal{D}, a \sim \pi_\theta(a|s)} \left[Q_\phi(s, a)\right] - \beta \mathbb{D}_{\rm KL} \left[ \pi_\theta (a|s) || \pi^{i-1} (a|s) \right]
    \end{equation}
\EndFor
\State \Return Best $\pi \in \{\pi_1, \dots, \pi_K\}$ on validation dataset
\end{algorithmic}
\end{algorithm}

\paragraph{Stage 2: Train Process Reward Model.}
At iteration $i$, the PRM $Q_i$ is trained via supervised learning on dataset $\mathcal{D}$. We use a soft binary cross-entropy (BCE) loss, treating $\hat{Q}(s, a)$ as a soft label:
\begin{equation}
    \mathcal{L}(Q_\phi) = -\mathbb{E}_{(s, a, \hat{Q}) \sim \mathcal{D}} \left[ \hat{Q} \log Q_\phi(s, a) + (1 - \hat{Q}) \log (1 - Q_\phi(s, a)) \right].
\end{equation}
The PRM is updated by minimizing this loss $Q^i = \arg\min_{Q_\phi} \mathcal{L}(Q_\phi)$. Note that this stage is similar to training a reward model in RLHF, where the loss function is a Bradley-Terry (BT) loss on preference data. We too explore using a BT loss as an ablation in Sec.~\ref{sec:agent_prm:exp}.

\paragraph{Stage 3: Train Policy via RL.}
Finally, we update the policy $\pi_i$ to maximize the PRM while staying close to the previous policy.
\begin{equation}
\pi_{i} = \arg\max_{\pi_\theta} \mathbb{E}_{s \sim \mathcal{D}, a \sim \pi_\theta(a|s)} \left[Q_\phi(s, a)\right] - \beta \mathbb{D}_{\rm KL} \left[ \pi_\theta (a|s) || \pi^{i-1} (a|s) \right]
\end{equation}
The above can be solved via standard RLHF frameworks that employ PPO~\cite{schulman2017proximal}, Online DPO~\cite{guo2024direct}, or Rejection Sampling~\cite{dubey2024llama}. We use Online DPO in our experiments. 

Notably, the policy is regularized to stay close to $\pi_{i-1}$ rather than the initial SFT policy. Since the PRM is trained on rollouts generated by $\pi_{i-1}$, straying too far from this reference can degrade PRM accuracy. This aligns with the principle of conservative policy iteration~\cite{kakade2002approximately}, where policies are updated within a restricted distributional shift to maintain validity of learned reward estimates. This approach is also consistent with best practices in online DPO~\cite{guo2024direct}.

\paragraph{Inference.}  
At test time, we can improve policy execution using a Best-of-N strategy, denoted as $\mathrm{BoN}(\pi, Q)$. At each turn, we sample $N$ candidate responses from $\pi$ and select one with the highest PRM score $Q(s, a)$. This provides a simple yet effective way to leverage the process reward model for inference. 
Test-time scaling is controlled via $N$: increasing $N$ allows the agent to explore a wider set of responses while still relying on $Q$ for selection.

\subsection{Experiments}
\label{sec:agent_prm:exp}
\paragraph{Setup.}
We evaluate our approach on ALFWorld~\citep{shridhar2020alfworld}, a standard text-based game benchmark for language agents. Each task specifies a high-level goal, e.g., ``heat mug and put it in cabinet,'' which the agent must accomplish by issuing text commands (e.g., ``go to shelf 1,'' ``pick up mug 2''). Solving these tasks requires subgoal planning, progress tracking, and efficient object search (e.g., mugs are likely on shelves or in cabinets). 
Each task consists of $30$ timesteps. The dataset contains $6$ task categories, a training set of $3257$ games, and two evaluation sets: $139$ in-distribution tasks and $134$ out-of-distribution tasks. Performance is measured by task success rate (\texttt{\%suc}$\uparrow$) and average number of actions (\texttt{\#act}$\downarrow$).

We compare against a prior work BUTLER~\citep{shridhar2020alfworld} and a number of prompting baselines \texttt{ReAct}~\citep{yao2022react}, \texttt{Autogen gpt-3.5}~\citep{wu2023autogen}, \texttt{ExpeL gpt-3.5}~\citep{zhao2024expel}, \texttt{Reflexion gpt-3}~\citep{shinn2023reflexion} , \texttt{AdaPlanner gpt-3}~\citep{sun2024adaplanner}. The prompting baselines all use larger gpt models along with few-shot examples. Adaplanner and reflexion get multiple attempts on the same task at test time, which significantly boosts performance. We also add \react baselines using the exact same prompt that our fine-tuned agent uses, with  stronger models such as \gptfo\footnote{https://platform.openai.com/docs/models}, claude\footnote{https://docs.anthropic.com/en/docs/about-claude/models}, and gemini\footnote{https://ai.google.dev/gemini-api/docs/models/gemini}.

For \agentprm, we fine-tune Llama3.2-3B~\citep{dubey2024llama} for both PRM and policy models, and run the process for $3$ iterations. The policy $\pi_0$ is initialized using SFT data. At each iteration, we collect $10k$ rollout trajectories (parallelized) which are used to train the PRM and the generator. See code for hyperparameters and prompts for the agent. There are two modes of inference: using the policy $\pi_i$ directly or doing Best-of-N BoN($\pi$, $Q$) with policy $\pi$ and PRM $Q$ and $N=16$. 

\begin{table}[!t]
\centering
\begin{minipage}{\textwidth}
\centering
\resizebox{\textwidth}{!}{
\begin{tabular}{l|cc|c c c c c c}
\toprule
\multirow{2}{*}{Method} & \multicolumn{2}{c|}{\textbf{All tasks}} & \multicolumn{1}{c}{\textsc{Pick} tasks} & \multicolumn{1}{c}{\textsc{Clean} tasks} & \multicolumn{1}{c}{\textsc{Heat} tasks} & \multicolumn{1}{c}{\textsc{Cool} tasks} & \multicolumn{1}{c}{\textsc{Look} tasks} & \multicolumn{1}{c}{\textsc{Pick 2} tasks} \\
\cmidrule(lr){2-9}
& \texttt{\%suc}$\uparrow$ & \texttt{\#act}$\downarrow$ & \texttt{\%suc}$\uparrow$ & \texttt{\%suc}$\uparrow$  & \texttt{\%suc}$\uparrow$  & \texttt{\%suc}$\uparrow$  & \texttt{\%suc}$\uparrow$  & \texttt{\%suc}$\uparrow$ \\
\midrule
\texttt{BUTLER} \textcolor{perfred}{[1]}  & 35.0 & - & 50.0 & 74.0 & 83.0 & 91.0 & 39.0 & 65.0 \\
\texttt{ReAct} few-shot \textcolor{perfred}{[2]} & 57.0 & - & 65.0 & 39.0 & 83.0 & 76.0 & 55.0 & 24.0 \\
\texttt{Autogen gpt-3.5} \textcolor{perfred}{[3]}  & 77.0 & - & - & - & - & - & - & - \\
\texttt{ExpeL gpt-3.5} \textcolor{perfred}{[4]}  & 59.0 & - & - & - & - & - & - & - \\
\texttt{Reflexion gpt-3} \textcolor{perfred}{[5]}  & 88.0 & - & 75.0 & 90.3 & 91.3 & 90.5 & 88.9 & 94.1 \\
\texttt{AdaPlanner gpt-3} \textcolor{perfred}{[6]}  & 91.7 & - & 100.0 & 96.7 & 95.6 & 100.0 & 100.0 & 47.0 \\
\cmidrule(lr){1-9}
\texttt{ReAct} \gptfo & 65.7 & 20.2 & 91.7  & 35.5  & 56.5  & 52.4  & 100.0  & 76.5 \\
\texttt{ReAct} \gptfomini & 29.9 & 25.5 & 33.3  & 25.8  & 17.4  & 14.3  & 66.7  & 29.4 \\
\texttt{ReAct} \texttt{claude-3.5-sonnet} & 76.1 & 19.0 & \cellgreen 95.8  & 61.3  & 60.9  & 81.0  & 88.9  & 76.5 \\
\texttt{ReAct} \texttt{claude-3.5-haiku} & 16.4 & 27.2 & 33.3  & 9.7  & 8.7  & 9.5  & 38.9  & 0.0 \\
\texttt{ReAct} \texttt{gemini-1.5-flash} & 19.4 & 26.3 & 41.7  & 12.9  & 13.0  & 19.0  & 16.7  & 11.8 \\
\midrule
\texttt{Llama3.2-3B} $\pi_0$ & 64.9 & 14.9 & 62.5 & 74.2 & 69.6 & 71.4 & 66.7 & 35.3 \\
\texttt{Llama3.2-3B} BoN($\pi_0$, $Q_0$) & 67.9 & 15.1 & 66.7 & 74.2 & 69.6 & 71.4 & 66.7 & 52.9 \\
\texttt{Llama3.2-3B} $\pi_1$ & 73.9 & 14.0 & 58.3 & 80.6 & 73.9 & 71.4 & \cellgreen{100.0} & 58.8 \\
\texttt{Llama3.2-3B} BoN($\pi_1$, $Q_0$) & 84.3 & 13.5 & 75.0 & \cellgreen{90.3} & \cellgreen{95.7} & 76.2 & \cellgreen{100.0} & 64.7 \\
\texttt{Llama3.2-3B} $\pi_2$ & 85.8 & 12.6 & 75.0 & 87.1 & 91.3 & \cellgreen{100.0} & \cellgreen{100.0} & 58.8 \\
\texttt{Llama3.2-3B} BoN($\pi_2$, $Q_1$) & 88.8 & \cellgreen{12.0} & 79.2 & 87.1 & 91.3 & \cellgreen{100.0} & \cellgreen{100.0} & 76.5 \\
\texttt{Llama3.2-3B} $\pi_3$ & 88.1 & 12.7 & 79.2 & \cellgreen{90.3} & 91.3 & \cellgreen{100.0} & \cellgreen{100.0} & 64.7 \\
\texttt{Llama3.2-3B} BoN($\pi_3$, $Q_2$) & \cellgreen{91.0} & 12.5 & \cellgreen{87.5} & 87.1 & 91.3 & \cellgreen{100.0} & \cellgreen{100.0} & \cellgreen{82.4} \\

\bottomrule
\end{tabular}
}
\end{minipage}
\vspace{0.5em}
\caption{\small \textbf{\agentprm Evaluation on Alfworld} on $136$ out-of-distribution games (max $30$ actions). Baseline comparisons include \textcolor{perfred}{[1]} \texttt{BUTLER}~\citep{shridhar2020alfred}, \textcolor{perfred}{[2]} \texttt{ReAct} few-shot~\citep{yao2022react}, \textcolor{perfred}{[3]} \texttt{Autogen}~\citep{wu2023autogen}, \textcolor{perfred}{[4]} \texttt{ExpeL}~\citep{zhao2024expel}. Note  \textcolor{perfred}{[5]} \texttt{Reflexion}~\citep{shinn2023reflexion} and \textcolor{perfred}{[6]} \texttt{AdaPlanner}~\citep{sun2024adaplanner} make multiple attempts on the same test task, while we do not. We also add 
our own \textsc{ReAct} instruction prompt with different models. 
\agentprm with a 3B model across iterations ($\pi_1, \pi_2, \pi_3$) outperforms the stronger models like \texttt{claude-3.5-sonnet}. \figGap }
\label{tab:agent_prm_alfworld}
\end{table}

\begin{figure*}[t]
\centering
\includegraphics[width=\linewidth]{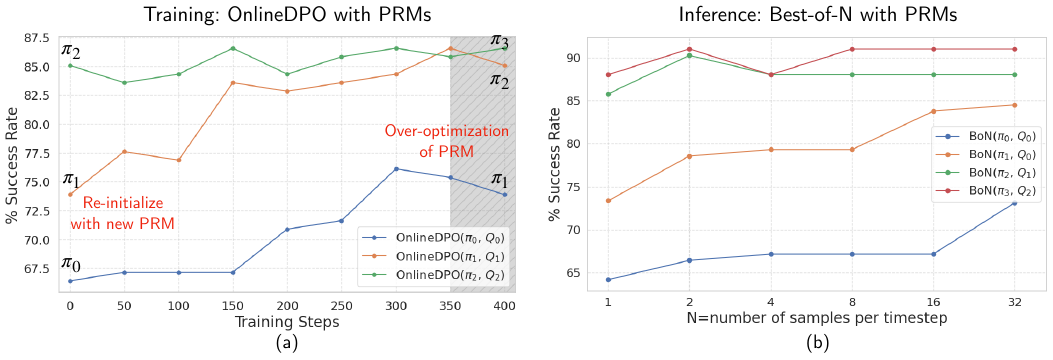}
\caption{\small \textbf{Training and Inference.} (a) Success rate vs training steps during online DPO with PRMs for $3$ iterations of \agentprm. $\pi_0$ is initialized with SFT. PRM $Q_0$ is trained on $\pi_0$ rollouts. OnlineDPO($\pi_0$, $Q_0$) is run for 400 training steps, during which the success rate goes up till it plateaus. The final checkpoint $\pi_1$ is taken and the process repeated to get $\pi_2, \pi_3$ till success rate limit is reached. 
(b) Inference with Best-of-N with varying $N=1, 2, \dots, 32$. For earlier policies $\pi_0, \pi_1$ success rate increases significantly, but scaling gains are limited for later policies $\pi_2, \pi_3$. \figGap \vspace{-1em}
}
\label{fig:agent_prm_training_inference}
\end{figure*}

\paragraph{Overall Results.}
Table~\ref{tab:agent_prm_alfworld} shows the performance of \agentprm against all baselines. \agentprm outperforms all baselines, with the best policy achieving $88.1\%$ success rate and $91.0\%$ success rate in best-of-N mode.\footnote{Adaplanner with gpt-3 has a higher success rate, but gets multiple attempts at test time rendering comparison unfair.} Iteration $2$ has the biggest performance gain $(73.9\% \rightarrow 85.8\%)$ \textbf{leading to a policy $\pi_2$ that surpasses the strongest model \texttt{claude-3.5-sonnet}} with higher success rate $(85.8\% > 76.1\%)$ and lower actions $(12.0 < 19.0)$. Best-of-N always leads to a higher performance gain, the iteration $1$ having the largest gain  $(73.9\% \rightarrow 84.3\%)$, eventually plateuing for iteration $3$ with $(88.1\% \rightarrow 91.0\%)$. 

% Interestingly, for the \textsc{Pick 2} task, the Best-of-N has a persistent gain, indicating a potential generator-verifier gap for that subtask.

\paragraph{Training Curves.}
Fig.~\ref{fig:agent_prm_training_inference} (a) shows how success rate evolves during policy training via RL (Stage 3). Success improves across iterations ($\pi_0$: 64.9\%, $\pi_1$: 73.9\%, $\pi_2$: 85.8\%,  $\pi_3$: 88.1\%), with each policy achieving higher success than its predecessor. 
At each iteration, success rate increases over training steps but eventually plateaus due to over-optimization—i.e., the policy exploits the PRM beyond its training distribution. Re-training with the updated PRM mitigates this issue and enables further improvements, though performance saturates at $\pi_3$, likely due to model capacity limits.  
The largest improvement occurs between $\pi_1$ ($73.9\%$) and $\pi_2$ ($85.8\%$), with gains appearing early in training (within 150 steps). In contrast, $\pi_0 \rightarrow \pi_1$ gains emerge later (after 150 steps). This suggests that $Q_1$ is trained on more successful trajectories than $Q_0$, providing a better optimization landscape for policy improvement.

\paragraph{Test-time Scaling.}
Fig.~\ref{fig:agent_prm_training_inference} (b) shows success rates in Best-of-N mode as $N$ varies from $1$ to $32$. For earlier policies ($\pi_0, \pi_1$), performance improves significantly as $N$ increases, with the largest gains for $N > 16$. However, for later policies ($\pi_2, \pi_3$), scaling gains diminish. This is both due to the limited head-room, but also due to reward over-optimization which we discuss next. 

\paragraph{Question: Can we measure and mitigate reward hacking?}

\begin{wrapfigure}{r}{0.5\textwidth} % 'r' for right alignment, 0.5\textwidth for half page width
    \centering
    \includegraphics[width=0.48\textwidth]{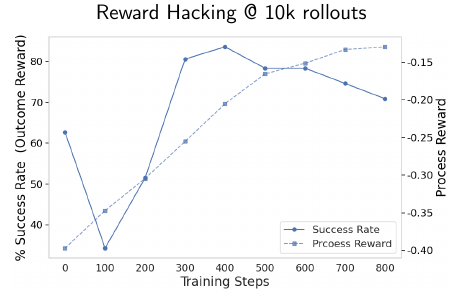} % Adjust the width as needed
    \caption{\small \textbf{Process Reward Hacking.} Success rate (outcome reward) and process reward over training step for a PRM trained with 10k rollouts. Process reward on validation data keeps increasing while outcome reward peaks and then degrades. \figGap \vspace{-2em}}
    \label{fig:reward_hacking}
\end{wrapfigure}

A common issue in RLHF-style training is \emph{reward hacking}~\cite{krakovna2020specification, weng2024rewardhacking}, where the policy optimizes the learned reward model rather than achieving true task success. This occurs when:
\begin{enumerate}[leftmargin=0.2in, nosep]
    \item The policy drifts too far from the distribution on which the PRM was trained.
    \item The PRM is trained on insufficient rollouts, leading to poor generalization.
\end{enumerate}
We control for (1) and investigate (2) by training PRMs on $10k$ vs. $70k$ rollouts.

Fig.~\ref{fig:reward_hacking} shows how both success rates (outcome rewards) and process rewards vary over training steps when the PRM is trained over 10k rollouts. After $400$ steps, the success rate begins to fall from $82\%$ to $70\%$. In contrast, the reward on the validation set keeps increasing. This shows clear signs of reward hacking. An open question remains how to reliably detect over-optimization without evaluating the success rate (which is difficult to scale). We tried an ensemble technique, training multiple reward models on different partitions of the data, but they all increased over training steps. 
% Mitigation steps are to either (1) increase the number of rollouts in the PRM (2) increase the $\beta$ to regularize further.

\paragraph{Question: Can we train PRMs on relative vs absolute losses?}
While we train PRMs using an absolute fashion, i.e., predict $Q(s,a)$, we use them in a relative fashion:
(1) During training (online DPO), the PRM ranks two different responses by the policy.
(2) During inference, the PRM ranks different responses generated by the policy.
This raises the question: should PRMs predict absolute values ($Q^\pi(s,a)$) or relative values ($A^\pi(s,a)$)? 

From an RL perspective, advantage functions $A^\pi(s,a) = Q^\pi(s,a) - V^\pi(s)$ often exhibit lower variance, improving stability during training. Prior work in mathematical reasoning~\cite{setlur2024rewarding} has also made similar arguments for training PRMs as advantage estimators. Intuitively, it might be difficult to judge how good an action in a globally normalized manner, but much easier to judge the action locally among other functions. 

\begin{wrapfigure}{r}{0.5\textwidth} % 'r' for right alignment, 0.5\textwidth for half page width
    \centering
    \includegraphics[width=0.48\textwidth]{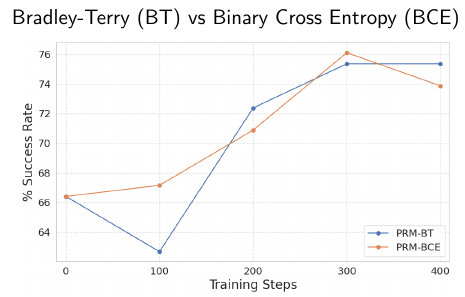} % Adjust the width as needed
    \caption{ \small \textbf{Absolute vs Relative Loss for PRM}. Success rate over training steps for PRM trained with $70k$ rollouts. Both losses lead to similar performance. \figGap \vspace{-1em}}
    \label{fig:relative_absolute_prm}
\end{wrapfigure}

To train PRMs in a relative manner, we use the following procedure:
\begin{enumerate}[leftmargin=0.2in, nosep]
    \item \textbf{(Stage 1)} Collect rollouts and construct a dictionary $\mathcal{G}(s)$ that maps each state to its sampled actions and corresponding $Q$ values.
    \item \textbf{(Stage 1)} Construct a preference dataset consisting of ranked action pairs $(s, a_1 \geq a_2)$, where $Q(s,a_1) - Q(s,a_2) \geq \delta$. Here, $\delta$ is a hyperparameter that defines a minimum margin for preference.
    \item \textbf{(Stage 2)} Train $Q$ using a Bradley-Terry loss~\cite{bradley1952rank}:  
    $-\mathbb{E}_{(s, a_1, a_2) \sim \mathcal{D}} [\log \sigma(Q_\phi(s, a_1) - Q_\phi(s, a_2))]$
    where $\sigma()$ is the sigmoid function.
\end{enumerate}

Fig.~\ref{fig:relative_absolute_prm} compares PRMs trained with absolute vs. relative losses. Surprisingly, both approaches yield similar performance. One explanation is that the dataset size for absolute vs relative is not equal. If a state isn't visited multiple times, it is discarded for the relative loss. There are far fewer states that are visited multiple times, leading to a small dataset and hence higher error for the PRM. 

\section{Inverse Process Reward Models}
\label{sec:inverse_prm}
The agent PRM framework in Sec.~\ref{sec:agent_prm} assumes access to outcome rewards, which may not always be available. Designing rewards manually is labor-intensive and susceptible to misspecification~\cite{krakovna2020specification, pan2022effects}, as it requires explicitly capturing every success and failure condition. Instead, consider a setting where the agent has access only to expert demonstrations—sequences of successful actions performed by a human, rule-based agent, or promoted LLM agent. The key challenge is: \emph{How can we learn process reward models solely from demonstrations, without access to explicit outcome rewards?}

\subsection{Formulation}
Given a set of expert demonstrations $\mathcal{D}^* = \{(s^\star, a^\star)\}$, the goal is to infer a reward function $r(s, a)$\footnote{Note this is a one-step reward, different from process rewards which are Q-values, i.e., cumulative rewards.} that explains expert behavior. We formulate this as inverse reinforcement learning (IRL), which learns a reward that maximizes the expert's expected return relative to any other policy. Formally, IRL can be posed as a min-max adversarial game between a reward player $r(s,a)$ (discriminator) and a policy player $\pi$  (generator):
\begin{equation}
\label{eq:irl_game}
\min_{\pi} \max_r \mathbb{E}_{\pi^*} \left[r(s^\star, a^\star)\right] - \mathbb{E}_{\pi} [r(s, a)].
\end{equation}

This game is solved iteratively. At each iteration $i$, the reward function $r_i(s,a)$ is updated to distinguish expert demonstrations from \emph{all past} learner policies (no-regret update). The policy player $\pi_i(a|s)$ then optimizes against the updated reward function (best response update):
\begin{equation}
\label{eq:irl_update}
\begin{tikzpicture}[>=latex, node distance=1cm]
  % Left node with the first equation
  \node (left) {
    $\displaystyle r_i = \arg\max_{r}\, \mathbb{E}_{\pi^*}\Bigl[r(s^\star,a^\star)\Bigr] - \mathbb{E}_{\pi_{0:i-1}}\Bigl[r(s,a)\Bigr]$
  };

  % Right node with the second equation, placed 2cm to the right of the left node
  \node (right) [right=2cm of left] {
    $\displaystyle \pi_i = \arg\max_{\pi}\, \mathbb{E}_{\pi}\Bigl[r_i(s,a)\Bigr]$
  };

  % Top arrow from the top-right of the left node to the top-left of the right node
  \draw[->, bend left=20] (left.north east) to (right.north west);
  
  % Bottom arrow from the bottom-left of the right node to the bottom-right of the left node
  \draw[->, bend left=20] (right.south west) to (left.south east);
\end{tikzpicture}
\end{equation}
where sampling from $\pi_{0:i-1}$ amounts to aggregating $(s,a)$ data from all past policies and sampling uniformly from that. 

\paragraph{IRL via PRMs.}
A naive IRL implementation would require an outer optimization loop around the agent PRM framework, making it computationally impractical. Instead, we use a telescoping identity to express the one-step reward in terms of Q-values, allowing direct estimation of the PRM. Specifically, we rewrite the reward function as:
\footnote{This identity holds for any Q-function, but we use $Q^\pi$ since we can sample on-policy.}
\begin{equation}
\label{eq:rasq}
r(s, a) = Q^\pi(s, a) - \gamma \mathbb{E}_{a' \sim \pi} Q^\pi(s', a').
\end{equation}

Writing the reward in terms of Q, or the verifier in terms of a generator, is an age-old trick that has been used effectively in various imitation learning~\cite{garg2021iq} and reinforcement learning formulations~\cite{xie2020q}.

We revisit the IRL update (\ref{eq:irl_update}) but replace the one-step reward with the PRM parameterization in (\ref{eq:rasq}). At iteration $i$, the update for PRM $Q_i^\pi$ is:
\begin{equation}
\begin{aligned}
    Q_i^\pi  = \arg\max_{Q} \quad &  \mathbb{E}_{ \substack{(s^\star,a^\star,s'^\star)\sim\pi^* \\ a' \sim \pi_{i-1}(.|s'^\star)}} \left[ Q(s^\star, a^\star) - \gamma Q(s'^\star, a') \right] - \\ 
    &  \mathbb{E}_{ \substack{(s,a,s')\sim\pi_{0:i-1} \\ a' \sim \pi_{i-1}(.|s')}}  [Q(s, a) - \gamma Q(s', a')]
\end{aligned}
\end{equation}
Here, the difference in Q-values increases along expert trajectories $(s^\star,a^\star,s'^\star)$ and decreases along all past learner trajectories $(s,a,s')$. Since $Q_i^\pi$ estimates the Q-values for the current policy $\pi_{i-1}$, the action $a'$ is always sampled from $\pi_{i-1}$.

The policy update remains an RL step, where $\pi_i$ is trained to maximize the learned PRM, following the same procedure as in Sec.~\ref{sec:agent_prm}:
\begin{equation}
    \pi_i = \arg\max_\pi \mathbb{E}_{\pi} [Q_i^\pi(s, a)] - \beta \mathbb{D}_{\rm KL} \left[ \pi (a|s) \; || \; \pi^{i-1} (a|s) \right]
\end{equation}

\subsection{Approach}
Algorithm~\ref{alg:inverse_prm} describes \inverseprm: a simple three-stage iterative process to learn and refine PRMs and policies given expert demonstration. 

\begin{enumerate}[leftmargin=0.2in, nosep]
\item Create positive $\mathcal{D}^+$ and negative $\mathcal{D}^-$ transitions using expert demos and rollouts from $\pi_{i-1}$.
\item Train the PRM $Q_i(s,a)$ to discriminate between $\mathcal{D}^+$ and $\mathcal{D}^-$ (\emph{similar to RLHF})
\item Train the policy $\pi_i$ using reinforcement learning on the trained RM (\emph{similar to RLHF})
\end{enumerate}  
The framework is very similar to the three stage process in \agentprm (Algorithm~\ref{alg:prm_training}) with the difference being no outcome reward and instead expert demonstrations. Stage 1 and 2 differ to accommodate this, while Stage 3 remains the same. Just like \agentprm, the algorithm for \inverseprm builds on existing RLHF frameworks making it easy to implement and use. We describe each stage in detail below:

\begin{algorithm}[!t]
\caption{Inverse PRM}
\label{alg:inverse_prm}
\begin{algorithmic}[1]
\State \textbf{Initialize:} Policy $\pi_0$, expert demonstrations $\mathcal{D}^+ = \{ (s^*, a^*, s'^*) \}$, negative dataset $\mathcal{D}^- = \{\}$
\For{iteration $i = 1, \dots, K$}
    \Statex \Comment{\textbf{Stage 1: Construct Positive and Negative Transitions}}
    \State Collect rollouts $\mathcal{D}_i = \{ (s, a, s', a') \}$ using policy $\pi_{i-1}$
    \State Aggregate into the negative dataset: $\mathcal{D}^- \gets \mathcal{D}^- \cup \mathcal{D}_{i}$
    \State Relabel next actions: $a' \sim \pi_{i-1}(s')$ for all $(s, a, s', a') \in \mathcal{D}^- \cup \mathcal{D}^+$
    
    \Statex \Comment{\textbf{Stage 2: Train Process Reward Model}}
    \State Train PRM $Q_i$ by minimizing the classification loss:
    \begin{equation*}
        \begin{aligned}
            \mathcal{L}(\phi) = &-\mathbb{E}_{(s^*, a^*, s'^*, a') \sim \mathcal{D}^+} \left[ \log \sigma(Q_\phi(s^*, a^*) - \gamma Q_\phi(s'^*, a')) \right] \\
            &+\mathbb{E}_{(s, a, s', a') \sim \mathcal{D}^-} \left[ \log (1 - \sigma(Q_\phi(s, a) - \gamma Q_\phi(s', a'))) \right]
        \end{aligned}
    \end{equation*}

    \Statex \Comment{\textbf{Stage 3: Train Policy via RL}}
    \State Update policy $\pi_i$ to maximize $Q_i$ while regularizing to $\pi_{i-1}$:
    \begin{equation}
    \pi_{i} = \arg\max_{\pi_\theta} \mathbb{E}_{s \sim \mathcal{D}_i, a \sim \pi_\theta(a|s)} \left[Q_\phi(s, a)\right] - \beta \mathbb{D}_{\rm KL} \left[ \pi_\theta (a|s) || \pi^{i-1} (a|s) \right]
    \end{equation}
\EndFor
\State Best $\pi \in \{\pi_1, \dots, \pi_K\}$ on validation dataset
\end{algorithmic}
\end{algorithm}

\paragraph{Stage 1: Create Positive / Negative Transitions.} 
We initialize with an positive dataset $\mathcal{D}^+ = \{(s^*, a^*, s'^*)\}$ containing state, action, next-state transitions from expert demonstrations. At iteration $i$, we rollout policy $\pi_{i-1}$ in the environment to collect  $\mathcal{D}_i = \{(s, a, s',a')\}$ to get state, action, next-state, next-action transitions. These rollouts are then aggregated with an existing negative dataset $\mathcal{D}^- \gets \mathcal{D}^- \cup \mathcal{D}_{i}$. Finally, the next-action in both $\mathcal{D}^+$ and $\mathcal{D}^-$ are relabeled by calling $a' \sim \pi_{i-1}(s')$. We end up with a positive dataset $\mathcal{D}^+ = \{(s^*, a^*, s'^*, a')\}$ where the transitions are from expert demonstrations, and negative dataset  $\mathcal{D}^- = \{(s, a, s', a')\}$ where the transitions are from all previous learner policies. 

\paragraph{Stage 2: Training Process Reward Model.}
At iteration $i$, the PRM $Q_i(s, a)$ is trained to distinguish expert transitions $\mathcal{D}^+$ from learner transitions $\mathcal{D}^-$. We frame this as a binary classification problem, where expert transitions are labeled as positive (1) and learner transitions as negative (0). 

A key distinction from standard reward modeling is that the classifier operates on the \emph{difference of PRM values}, $Q_\phi(s, a) - \gamma Q_\phi(s', a')$, capturing the relative advantage of one transition over another. The loss function is:
\begin{equation*}
    \begin{aligned}
        \mathcal{L}(\phi) = &-\mathbb{E}_{(s^*, a^*, s'^*, a') \sim \mathcal{D}^+} \left[ \log \sigma(Q_\phi(s^*, a^*) - \gamma Q_\phi(s'^*, a')) \right] \\
        &+\mathbb{E}_{(s, a, s', a') \sim \mathcal{D}^-} \left[ \log (1 - \sigma(Q_\phi(s, a) - \gamma Q_\phi(s', a'))) \right]
    \end{aligned}
\end{equation*}

\paragraph{Stage 3: Train Policy via RL.}
The policy update follows the same procedure as in \agentprm: the policy $\pi_i$ is optimized to maximize the PRM $Q_i$ while remaining close to the previous iteration's policy $\pi_{i-1}$. Formally, we solve:
\begin{equation}
\pi_{i} = \arg\max_{\pi_\theta} \mathbb{E}_{s \sim \mathcal{D}, a \sim \pi_\theta(a|s)} \left[Q_\phi(s, a)\right] - \beta \mathbb{D}_{\rm KL} \left[ \pi_\theta (a|s) || \pi^{i-1} (a|s) \right]
\end{equation}
As in \agentprm, the KL regularization ensures stability by preventing $\pi_i$ from straying too far from the reference policy, mitigating distribution shift and reward hacking risks.

\subsection{Experiments}
\label{sec:inverse_prm:exp}
\paragraph{Setup.}
We evaluate \inverseprm using an expert policy from our prior work, \texttt{LEAP}~\citep{choudhury2024better}, a Llama-3-8B model trained via privileged feedback from \gptfo. We sample $10k$ expert demonstrations and train \inverseprm for $2$ iterations. The policy $\pi_0$ is initialized identically to \agentprm. At each iteration, we collect rollouts to ensure the aggregated negative dataset contains $10k$ trajectories. As in \agentprm, inference can be performed directly using the trained policy or via Best-of-N selection $\mathrm{BoN}(\pi, Q)$. See code for hyperparameters and agent prompts.

We compare \inverseprm against two baselines: (1) SFT: A policy trained directly on expert demonstrations. (2) \agentprm: A policy trained using only outcome rewards, without expert demonstrations, but with increased rollouts ($70k$ rollouts).

\paragraph{Overall Results.}
Table~\ref{tab:inverse_prm_alfworld} compares \inverseprm with SFT and \agentprm. \inverseprm outperforms both baselines, with its final policy $\pi_2$ approaching expert performance ($86.6 \% < 91.0 \%$). \inverseprm significantly outperforms SFT on the same expert demonstrations ($86.6 \% > 63.4 \%$). The key reason is that SFT policies struggle to recover once they deviate from expert trajectories, whereas \inverseprm actively interacts with the environment to correct mistakes. Compared to \agentprm trained with $70k$ rollouts, \inverseprm achieves \emph{substantial gains in just one iteration} ($82.8 \% > 73.9 \%$). This highlights that leveraging dense expert demonstrations enables far greater sample efficiency than training purely with outcome rewards.

\begin{table}[!t]
\centering
\begin{minipage}{\textwidth}
\centering
\resizebox{\textwidth}{!}{
\begin{tabular}{l|cc|c c c c c c}
\toprule
\multirow{2}{*}{Method} & \multicolumn{2}{c|}{\textbf{All tasks}} & \multicolumn{1}{c}{\textsc{Pick} tasks} & \multicolumn{1}{c}{\textsc{Clean} tasks} & \multicolumn{1}{c}{\textsc{Heat} tasks} & \multicolumn{1}{c}{\textsc{Cool} tasks} & \multicolumn{1}{c}{\textsc{Look} tasks} & \multicolumn{1}{c}{\textsc{Pick 2} tasks} \\
\cmidrule(lr){2-9}
& \texttt{\%suc}$\uparrow$ & \texttt{\#act}$\downarrow$ & \texttt{\%suc}$\uparrow$ & \texttt{\%suc}$\uparrow$  & \texttt{\%suc}$\uparrow$  & \texttt{\%suc}$\uparrow$  & \texttt{\%suc}$\uparrow$  & \texttt{\%suc}$\uparrow$ \\
\midrule
Expert Policy* & 91.0 &  11.9 & 83.3  & 90.3  &  91.3  &  95.2  &  94.4  &  94.1 \\
\midrule
SFT & 63.4 & 13.9 & 79.2 & 80.6 & 69.6 & 52.4 & 50.0 & 29.4 \\
\midrule
\agentprm $\pi_0$ & 64.9 & 14.9 & 62.5 & 74.2 & 69.6 & 71.4 & 66.7 & 35.3 \\
\agentprm $\pi_1$ & 73.9 & 14.0 & 58.3 & 80.6 & 73.9 & 71.4 & \cellgreen{100.0} & 58.8 \\
\agentprm $\pi_2$ & 85.8 & 12.6 & 75.0 & 87.1 & \cellgreen 91.3 & \cellgreen{100.0} & \cellgreen{100.0} & 58.8 \\
\midrule
\inverseprm $\pi_0$ & 64.9 & 14.9 & 62.5 & 74.2 & 69.6 & 71.4 & 66.7 & 35.3 \\
\inverseprm  $\pi_1$ & 82.8 & 13.1 & \cellgreen{83.3} & \cellgreen{96.8} & 73.9 & 95.2 & \cellgreen{100.0} & 35.3 \\
\inverseprm $\pi_2$ & \cellgreen{86.6} & \cellgreen{12.5} & 79.2 & 90.3 & \cellgreen{91.3} & \cellgreen{100.0} & 94.4 & \cellgreen{64.7} \\
\bottomrule
\end{tabular}
}
\end{minipage}
\vspace{0.5em}
\caption{\small \textbf{Evaluation of \inverseprm on ALFWorld.} Success rates (\%) on 136 out-of-distribution tasks (max 30 actions). \inverseprm is trained on 10K expert demonstrations over 2 iterations. It outperforms SFT on expert demonstrations (86.6\% vs. 63.4\%). Compared to \agentprm trained with 70K rollouts, \inverseprm achieves a significantly higher success rate in iteration 1 (82.8\% vs. 73.9\%) and approaches expert-level performance (86.6\% vs. 91.0\%). By leveraging dense expert demonstrations, \inverseprm achieves greater sample efficiency than \agentprm. \figGap }
\label{tab:inverse_prm_alfworld}
\end{table}
\begin{figure*}[!t]
\centering
\includegraphics[width=\linewidth]{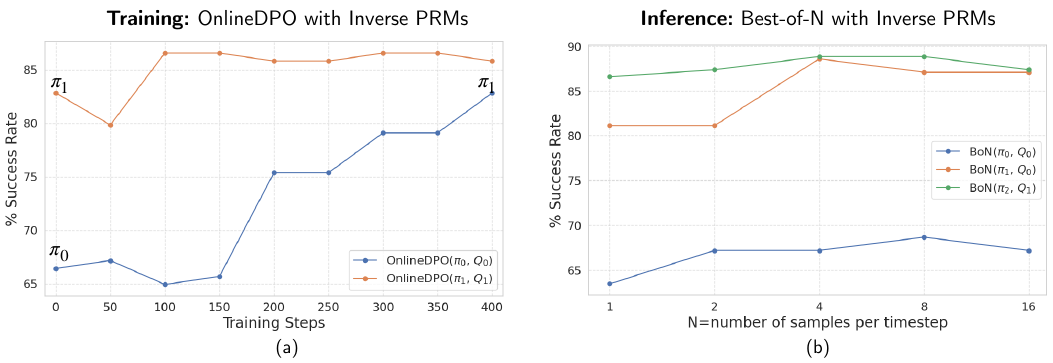}
\caption{
\small \textbf{Training and Inference of \inverseprm.} (a) Success rate (\%) vs. training steps for 2 iterations of \inverseprm using online DPO with PRMs. The initial policy $\pi_0$ is initialized identically to \agentprm. PRM $Q_0$ is trained on $\pi_0$ rollouts. $\mathrm{OnlineDPO}(\pi_0, Q_0)$ runs for 400 training steps, where success rate increases to near peak performance before saturating in iteration 2.  
(b) Best-of-N inference results for varying $N = \{1, 2, \dots, 32\}$. Policy quality has a greater impact than the PRM or $N$: $\mathrm{BoN}(\pi_0, Q_0)$ provides only modest improvement (64.9\% $\rightarrow$ 69.0\%), whereas $\mathrm{BoN}(\pi_1, Q_0)$ reaches 88.0\%. Performance saturates in iteration 2 ($\mathrm{BoN}(\pi_2, Q_1)$).  \figGap 
}
\label{fig:tasks}
\end{figure*}

\paragraph{Training Curves.}
Fig.~\ref{fig:agent_prm_training_inference} (a) shows the success rate evolution during policy training (Stage 3). The success rate improves dramatically in the first iteration ($64.9 \% \rightarrow 82.8 \%$), whereas \agentprm required multiple iterations to reach similar performance. 
This difference arises from the \emph{exploration challenge}~\citep{swamy2023inverse}: \agentprm must discover high-reward actions through trial-and-error, whereas \inverseprm benefits from expert demonstrations that implicitly capture successful strategies. We further analyze these exploration advantages in later sections.

\paragraph{Test-time Scaling.}
Fig.~\ref{fig:agent_prm_training_inference} (b) shows the effect of Best-of-N sampling on success rate as $N$ varies from $1$ to $32$. The policy quality has a greater impact than scaling $N$. For instance, increasing $N$ provides only moderate gains for $\mathrm{BoN}(\pi_0, Q_0)$ ($64.9 \% \rightarrow 69.0 \%$), but has a much larger effect for $\mathrm{BoN}(\pi_1, Q_0)$ ($88.0 \%$). Performance saturates with $\mathrm{BoN}(\pi_2, Q_1)$.

\section{Challenges and Opportunities}
Reinforcement learning presents several challenges, some well-known in RL (e.g., exploration) and others specific to LLM agents (e.g., model-predictive reasoning). Addressing these challenges requires both established RL/IL techniques—such as reset distributions and reward shaping—and novel strategies leveraging LLM-specific capabilities, such as steered exploration. 

\subsection{Exploration}
\label{sec:challenge:exploration}
\begin{wrapfigure}{r}{0.5\textwidth}
    \centering
    \includegraphics[width=0.48\textwidth]{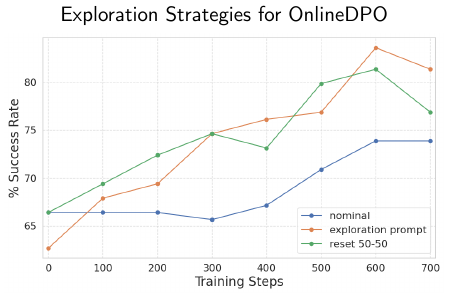} 
    \caption{\small \textbf{Different exploration strategies.} Success rate vs training steps with $\mathrm{OnlineDPO}(\pi_0, Q_0)$. Both \texttt{Reset-50-50} and \texttt{SteeredExploration} learn faster and reach higher performance. \figGap \vspace{-1.em}}
    \label{fig:exploration}
\end{wrapfigure}

Exploration remains a fundamental challenge in RL, requiring agents to explore effectively at both the \textit{turn level} (solving multi-step tasks) and the \textit{token level} (generating improved reasoning and actions). 
Fig.~\ref{fig:exploration} shows that the first iteration of \agentprm progresses slowly, requiring over $500$ training steps before ramping up and plateauing at $73.9\%$ success rate.

Traditional exploration strategies include stochastic action selection methods such as \textit{$\epsilon$-greedy}, entropy bonuses, or adjusting sampling temperature. However, these approaches do not scale well to high-dimensional, long-horizon tasks where reasoning quality is crucial. Instead, we explore structured strategies that leverage LLM-specific capabilities to guide exploration.

\paragraph{Strategy 1: Reset Distribution.}
A simple yet effective exploration strategy is to reset the agent to a \emph{good} distribution of states $\rho(s)$ that an optimal policy is likely to visit. A good distribution is one that covers optimal state distribution\footnote{Formally, a bounded density ratio $|\frac{\rho(s)}{d^{\pi^\star}(s)}| \leq C$, see \citep{bagnell2003policy}.}. Practitioners often use a $50\%-50\%$ reset distribution~\cite{swamy2023inverse}, where $50\%$ of initial states are sampled from successful expert demonstrations—such as human demonstrations or rule-based policies—while the remaining $50\%$ come from the agent’s on-policy rollouts. Intuitively, this approach helps bootstrap learning by exposing the agent to good states early, making it easier to recover from errors. We call this strategy \texttt{Reset-50-50}.

Fig.~\ref{fig:exploration} shows \texttt{Reset-50-50} for $\mathrm{OnlineDPO}(\pi_0, Q_0)$ where the distribution of states (prompts) is a mixture of $50\%$ states visited by $\pi_0$ and $50\%$ states visited by the expert policy from Sec.~\ref{sec:inverse_prm:exp}. Note the only change is the set of prompts used in Stage 3, everything else including the starting policy and PRM remains the same. We observe\texttt{Reset-50-50} learns much faster and reaches a higher peak $82\% > 73.9\%$. By simply exposing the policy to good states and optimizing the same PRM, the policy learns to generate improved reason-action that helps the policy recover from other states. 

\paragraph{Strategy 2: Steered Exploration.}
Unlike conventional RL policies, LLMs can be explicitly \emph{prompted} to explore, rather than relying on stochastic action selection. We call this strategy \texttt{Steered Exploration}. Concretely, during RL (stage 3), we inject a small addition to the agent prompt:
\begin{lstlisting}[language=, caption=Injected prompt snippet in \texttt{Steered Exploration}]
Use the following strategy for generating actions:
*  In your REASON, try to come up with a strategy for how you want to solve the task. This strategy could be a hypothesis of where the object might be based on your history of observations. Then base your ACTION on the REASON.
* Try to explore possible strategies
\end{lstlisting}
We remove this addition while training the agent, i.e., the agent still trains on the original prompt. This results in the generation of reason-actions that are more diverse than sampling reason-actions, but are of a much higher quality than simply increasing the temperature. 

Fig.~\ref{fig:exploration} shows the \texttt{Steered Exploration} strategy for $\mathrm{OnlineDPO}(\pi_0, Q_0)$. Again, the only thing that changed is how we are sampling reason-actions in Stage 3 (online DPO). We see that learning is much faster and reaches a much higher peak of $84\% > 73.9\%$. An explanation for why this works as well can be tied to Posterior Sampling for RL~\cite{osband2013more}; the LLM samples diverse ``models'' of how the world works (consistent with the history of observations) in its reason and proposes actions according to that model, while the PRM selects for the correct actions and consequently the correct model.

\paragraph{Strategy 3: Post-hoc Rationalization.}
The connection to posterior sampling yields another interesting way to do exploration. Suppose the agent had access to some privileged information, e.g., the future trajectory or hidden information about the MDP (hidden location of objects). Conditioned on that information, the agent can generate post-hoc rationalization for good actions to take. We explore training agents in this fashion in our prior work \texttt{LEAP}~\cite{choudhury2024better}. However, one challenge we faced is that not all post-hoc rationalizations are good, some are better than others.

Instead, we could imagine using this post-hoc rationalizer as an exploration policy. We call this strategy \texttt{PosteriorExplorer}. 
\texttt{PosteriorExplorer} suggests a diverse set of reason-actions that are then selected by the PRM based on which rationalization leads to good actions. The theory behind LEAP~\cite{swamy2022sequence, choudhury2018data} shows that the rationalizer learns a posterior over possible MDPs consistent with the POMDP the agent is solving, which is then refined by the RL procedure to select actions that lead to success. 

\subsection{Process Reward Shaping}
\label{sec:challenge:prm}
\begin{wrapfigure}{r}{0.5\textwidth} % 'r' for right alignment, 0.5\textwidth for half page width
    \centering
    \includegraphics[width=0.48\textwidth]{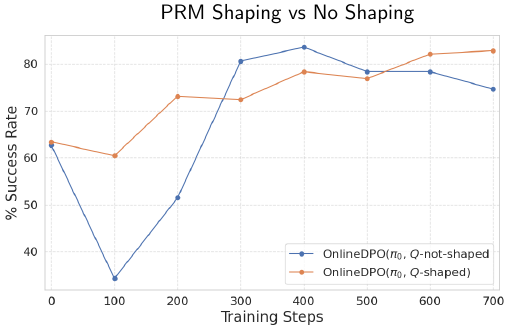} % Adjust the width as needed
    \caption{\small \textbf{Process Reward Shaping}. Success rate vs training steps of $\mathrm{OnlineDPO}(\pi_0, Q_0)$ when training with shaped rewards vs non-shaped rewards with $10k$ rollouts. Non-shaped rewards are noisy at low sample regimes with unstable performance. Shaped rewards lead to much more stable performance. \figGap \vspace{-1.em}}
    \label{fig:process_reward_shaping}
\end{wrapfigure}

Reinforcement learning from scratch is slow and sample inefficient. Practitioners often try to bootstrap RL using existing policies that have reasonable performance. 
We study a setting where only \textbf{10K rollout trajectories} can be collected, but a \textbf{reference policy} with moderate performance ($65.0\%$) is available.
We look at two such strategies: (1) initializing the agent via imitation learning and then doing RL, and (2) using \emph{process reward shaping}, where the reference policy provides structured guidance during RL training.

\paragraph{Strategy 1: Initialize with IL, then do RL.}
The simplest approach is to initialize the agent via SFT on trajectories generated by the reference agent. This ensures the initial policy is not random. 

Fig.~\ref{fig:process_reward_shaping} shows $\mathrm{OnlineDPO}(\pi_0, Q_0)$ for $10k$ rollouts where $\pi_0$ is initialized via SFT and then used for RL. We see that though $\pi_0$ begins at $64\%$, the training curve is unstable dropping to $32\%$ before climbing back up. Hence, even though the initialization is good, the policy unlearns some of that good behavior due to noise in the PRM. This would be true for more sophisticated imitation learning methods like DAGGER~\cite{ross2011reduction} because the reference policy is not used at all during RL.

\paragraph{Strategy 2: Process Reward Shaping.}
We next look at involving the reference policy in the RL process itself. We look at \emph{process reward shaping}, where instead of relying solely on sparse rewards, we shape the process reward using the advantage function of the reference policy. 

Given a reference policy $\mu$, we add a shaping term to the PRM target:
\begin{equation}
    Q(s, a) \leftarrow (1-\alpha) Q^\pi(s,a) + \alpha A^\mu(s, a)
\end{equation}
where $A^\mu(s, a)$ is the advantage w.r.t the reference policy $\mu$, i.e., $A^\mu(s, a) = r(s,a) + \gamma V^\mu(s') - V^\mu(s)$. 

$\alpha$ controls the power of the reference policy. Setting $\alpha=0$ recovers the original PRM. Setting $\alpha=1$ amounts to doing imitation learning, notable the AGGREVATE~\cite{ross2014reinforcement,sun2017deeply} algorithm. Our procedure is:
\begin{enumerate}[leftmargin=0.2in, nosep]
    \item Fit a value $V^\mu(s)$ using trajectories from the reference policy.
    \item In Stage 1, modify the PRM target to be $(1-\alpha) Q^\pi(s,a) + \alpha A^\mu(s, a)$
    \item Stage 2 and 3 remain unchanged. 
\end{enumerate}

Fig.~\ref{fig:process_reward_shaping} shows the shaped PRM training curves for $\alpha=0.5$. The learning is much more stable and continues to steadily rise to $700$ steps. This is because the $A^\mu(s, a)$, trained on much more rollouts from the reference policy ($70k$), counters the noisy PRM targets. Note that the learned policy significantly outperforms the reference policy ($82.0\% > 65.0 \%$), which IL alone would not have ensured. 

\subsection{Model-Predictive Reasoning}
\label{sec:challenge:model}
Recent large-scale RL advances have demonstrated promising results in multi-step reasoning tasks~\cite{guo2025deepseek}. However, applying RL to agentic settings remains challenging because each interaction requires querying the environment, significantly slowing down learning. This raises a key question: \textit{How can we reduce costly interactions while enabling agents to reason and plan effectively?}

One approach is to leverage learned world models. Instead of relying solely on trial-and-error, an LLM agent can simulate future trajectories using an internal model of the environment. This paradigm has been central in robotics, where real-world interactions are expensive and risky~\citep{abbeel2006application}. Model-based RL strategies, such as training policies in simulation before real-world deployment~\citep{andrychowicz2020learning}, have proven effective. Theoretically, generative models can provide mini-max optimal policies in model-based RL~\citep{agarwal2020model}.
We extend this perspective to LLM agents: \textit{Can we train them to plan (or deliberatively reason) with their internal models to improve decision-making?}

\paragraph{Strategy: Deliberative Reasoning with a Learned World Model.}
Instead of treating reasoning as a single-step process that immediately outputs an action, we propose a structured multi-stage approach where the agent explicitly predicts \textit{future consequences} before committing to an action. This decomposes the learning problem into two components: 
\begin{enumerate}[leftmargin=0.2in, nosep]
    \item \textbf{Learning a world model}: Train an internal reasoning model to predict future states given an action, using rollouts from the current agent.
    \item \textbf{Multi-turn planning and RL}: Optimize the agent’s reasoning process via reinforcement learning to maximize outcome rewards.
    \item \textbf{Plan-and-execute policy}: Structure the agent’s reasoning to first generate a complete plan, select the initial action, execute it, and then replan iteratively.
\end{enumerate}
This approach naturally connects to \textit{model-predictive control} (MPC), where agents reason over predicted trajectories before taking actions, rather than relying purely on reactive decision-making.

\vspace{-1em}
\section{Related Work}
\vspace{-1em}

\textbf{Fine-tuning agents.} 
Most of the work on LLM agents rely on prompting LLMs, e.g. ReAct~\citep{yao2022react}, Reflexion~\citep{shinn2023reflexion}, AdaPlanner~\citep{sun2024adaplanner}. However, prompting alone is insufficient to correct errors encountered at test-time~\citep{sodhi2024step, liu2023agentbench}. A simple way to improve LLMs is to fine-tune on successful trajectories generated manually or via a prompted LLM~\citep{schick2023toolformer, zeng2023agenttuning, chen2023fireact}. However, manually collecting demonstrations of reason and actions is challenging and hard to scale. 

Recent work \texttt{LEAP} has looked at leveraging privileged AI feedback~\citep{choudhury2024better} to design critics that distill the information into student agents, showing strong performance in text-based games, web navigation and interactive coding. However, the privileged correction in \texttt{LEAP} can be unrealizable for the agent, leading to poor success rates. Hence, we look at training agents directly using RL to maximize the outcome reward. 

Finally, ARCHER~\cite{zhou2024archer} proposes a very similar framework to train LLM agents using hierarchical RL. The Q-value is trained using temporal difference, while the policy is trained using REINFORCE. However, the results are limited to small models (GPT2). We simplify the framework so it connects with existing RLHF pipelines, do RL with Llama 3B models, propose novel algorithms like \inverseprm, and provide practical recipes like using reset distribution and reward shaping to improve efficiency. 

\textbf{Process Reward Models.} 
PRMs have mostly been looked at in the context of multi-stage math reasoning problems~\cite{cobbe2021training}, where they were trained in human annotation data to provide fine-grained supervision~\citep{lightman2023let, uesato2022solving}. Recent works look at automatically computing PRMs as Q value estimates~\citep{luo2024improve, wang2024math}. PRMs have been used to train generators~\citep{shao2024deepseekmath} and used for test-time scaling with beam search~\citep{snell2024scaling}, heuristic search~\cite{ma2023let} or tree search~\citep{wu2024inference}. 

There are interesting similarities and differences between PRMs used for math reasoning and the agent setting we look at here. Many works~\cite{havrilla2024teaching, shao2024deepseekmath, uesato2022solving} report small gains from optimizing PRMs rather than the outcome reward. In contrast, we see pretty strong gains with PRMs, where outcome reward is infeasible given long-horizons and limited access to the external environment. Some works have noted the reward-hacking / value-estimation issues with PRMs that we also analyze in Sec.~\ref{sec:agent_prm:exp}. To counter such issues, recent works~\citep{setlur2024rewarding} propose reward shaping PRMs using reference policies, which we also explore in Sec.~\ref{sec:challenge:prm}.

% SCORE

% Rewarding progress

\vspace{-1em}
\section{Conclusion}
\vspace{-1em}
We introduced \agentprm, a simple and scalable framework for training LLM agents using process reward models, and \inverseprm, which learns PRMs directly from demonstrations without explicit outcome rewards. Our results on ALFWorld show that small models trained with \agentprm outperform strong GPT-4o baselines, and \inverseprm achieves near-expert performance with significantly fewer rollouts. We outlined key challenges—exploration, process reward shaping, and model-predictive reasoning—and proposed methods that leverage both RL techniques and LLM-specific capabilities. Future work includes extending PRMs to richer agentic environments and exploring large-scale RL via model-predictive reasoning.

\bibliographystyle{unsrt} % Use unsrt to preserve citation order

\bibliography{llm_agents}

\begin{thebibliography}{10}

\bibitem{sodhi2024step}
Paloma Sodhi, SRK Branavan, Yoav Artzi, and Ryan McDonald.
\newblock Step: Stacked llm policies for web actions.
\newblock In {\em Conference on Language Modeling (COLM)}, 2024.

\bibitem{black2024pi_0}
Kevin Black, Noah Brown, Danny Driess, Adnan Esmail, Michael Equi, Chelsea Finn, Niccolo Fusai, Lachy Groom, Karol Hausman, Brian Ichter, et~al.
\newblock pi0: A vision-language-action flow model for general robot control.
\newblock {\em arXiv preprint arXiv:2410.24164}, 2024.

\bibitem{jimenez2023swe}
Carlos~E Jimenez, John Yang, Alexander Wettig, Shunyu Yao, Kexin Pei, Ofir Press, and Karthik Narasimhan.
\newblock Swe-bench: Can language models resolve real-world github issues?
\newblock {\em arXiv preprint arXiv:2310.06770}, 2023.

\bibitem{yao2022react}
Shunyu Yao, Jeffrey Zhao, Dian Yu, Nan Du, Izhak Shafran, Karthik Narasimhan, and Yuan Cao.
\newblock React: Synergizing reasoning and acting in language models.
\newblock {\em arXiv preprint arXiv:2210.03629}, 2022.

\bibitem{shinn2023reflexion}
Noah Shinn, Federico Cassano, Beck Labash, Ashwin Gopinath, Karthik Narasimhan, and Shunyu Yao.
\newblock Reflexion: Language agents with verbal reinforcement learning.(2023).
\newblock {\em arXiv preprint cs.AI/2303.11366}, 2023.

\bibitem{chen2023fireact}
Baian Chen, Chang Shu, Ehsan Shareghi, Nigel Collier, Karthik Narasimhan, and Shunyu Yao.
\newblock Fireact: Toward language agent fine-tuning, 2023.

\bibitem{khot2022decomposed}
Tushar Khot, Harsh Trivedi, Matthew Finlayson, Yao Fu, Kyle Richardson, Peter Clark, and Ashish Sabharwal.
\newblock Decomposed prompting: A modular approach for solving complex tasks.
\newblock {\em arXiv preprint arXiv:2210.02406}, 2022.

\bibitem{guo2025deepseek}
Daya Guo, Dejian Yang, Haowei Zhang, Junxiao Song, Ruoyu Zhang, Runxin Xu, Qihao Zhu, Shirong Ma, Peiyi Wang, Xiao Bi, et~al.
\newblock Deepseek-r1: Incentivizing reasoning capability in llms via reinforcement learning.
\newblock {\em arXiv preprint arXiv:2501.12948}, 2025.

\bibitem{haarnoja2018soft}
Tuomas Haarnoja, Aurick Zhou, Pieter Abbeel, and Sergey Levine.
\newblock Soft actor-critic: Off-policy maximum entropy deep reinforcement learning with a stochastic actor.
\newblock In {\em International conference on machine learning}, pages 1861--1870. PMLR, 2018.

\bibitem{lightman2023let}
Hunter Lightman, Vineet Kosaraju, Yura Burda, Harri Edwards, Bowen Baker, Teddy Lee, Jan Leike, John Schulman, Ilya Sutskever, and Karl Cobbe.
\newblock Let's verify step by step.
\newblock {\em arXiv preprint arXiv:2305.20050}, 2023.

\bibitem{uesato2022solving}
Jonathan Uesato, Nate Kushman, Ramana Kumar, Francis Song, Noah Siegel, Lisa Wang, Antonia Creswell, Geoffrey Irving, and Irina Higgins.
\newblock Solving math word problems with process-and outcome-based feedback.
\newblock {\em arXiv preprint arXiv:2211.14275}, 2022.

\bibitem{setlur2024rewarding}
Amrith Setlur, Chirag Nagpal, Adam Fisch, Xinyang Geng, Jacob Eisenstein, Rishabh Agarwal, Alekh Agarwal, Jonathan Berant, and Aviral Kumar.
\newblock Rewarding progress: Scaling automated process verifiers for llm reasoning.
\newblock {\em arXiv preprint arXiv:2410.08146}, 2024.

\bibitem{lambert2024tulu3}
Nathan Lambert, Jacob Morrison, Valentina Pyatkin, Shengyi Huang, Hamish Ivison, Faeze Brahman, Lester James~V. Miranda, Alisa Liu, Nouha Dziri, Shane Lyu, Yuling Gu, Saumya Malik, Victoria Graf, Jena~D. Hwang, Jiangjiang Yang, Ronan~Le Bras, Oyvind Tafjord, Chris Wilhelm, Luca Soldaini, Noah~A. Smith, Yizhong Wang, Pradeep Dasigi, and Hannaneh Hajishirzi.
\newblock Tülu 3: Pushing frontiers in open language model post-training.
\newblock 2024.

\bibitem{vonwerra2022trl}
Leandro von Werra, Younes Belkada, Lewis Tunstall, Edward Beeching, Tristan Thrush, Nathan Lambert, Shengyi Huang, Kashif Rasul, and Quentin Gallouédec.
\newblock Trl: Transformer reinforcement learning.
\newblock \url{https://github.com/huggingface/trl}, 2020.

\bibitem{shridhar2020alfred}
Mohit Shridhar, Jesse Thomason, Daniel Gordon, Yonatan Bisk, Winson Han, Roozbeh Mottaghi, Luke Zettlemoyer, and Dieter Fox.
\newblock Alfred: A benchmark for interpreting grounded instructions for everyday tasks.
\newblock In {\em Proc. {IEEE} Int. Conf. Computer Vision and Pattern Recognition}, 2020.

\bibitem{wang2024math}
Peiyi Wang, Lei Li, Zhihong Shao, Runxin Xu, Damai Dai, Yifei Li, Deli Chen, Yu~Wu, and Zhifang Sui.
\newblock Math-shepherd: Verify and reinforce llms step-by-step without human annotations.
\newblock In {\em Proceedings of the 62nd Annual Meeting of the Association for Computational Linguistics (Volume 1: Long Papers)}, pages 9426--9439, 2024.

\bibitem{snell2024scaling}
Charlie Snell, Jaehoon Lee, Kelvin Xu, and Aviral Kumar.
\newblock Scaling llm test-time compute optimally can be more effective than scaling model parameters.
\newblock {\em arXiv preprint arXiv:2408.03314}, 2024.

\bibitem{zheng2024sglang}
Lianmin Zheng, Liangsheng Yin, Zhiqiang Xie, Chuyue Sun, Jeff Huang, Cody~Hao Yu, Shiyi Cao, Christos Kozyrakis, Ion Stoica, Joseph~E Gonzalez, et~al.
\newblock Sglang: Efficient execution of structured language model programs.
\newblock {\em arXiv preprint arXiv:2312.07104}, 2024.

\bibitem{kwon2023efficient}
Woosuk Kwon, Zhuohan Li, Siyuan Zhuang, Ying Sheng, Lianmin Zheng, Cody~Hao Yu, Joseph~E. Gonzalez, Hao Zhang, and Ion Stoica.
\newblock Efficient memory management for large language model serving with pagedattention.
\newblock In {\em Proceedings of the ACM SIGOPS 29th Symposium on Operating Systems Principles}, 2023.

\bibitem{schulman2017proximal}
John Schulman, Filip Wolski, Prafulla Dhariwal, Alec Radford, and Oleg Klimov.
\newblock Proximal policy optimization algorithms.
\newblock {\em arXiv preprint arXiv:1707.06347}, 2017.

\bibitem{guo2024direct}
Shangmin Guo, Biao Zhang, Tianlin Liu, Tianqi Liu, Misha Khalman, Felipe Llinares, Alexandre Rame, Thomas Mesnard, Yao Zhao, Bilal Piot, et~al.
\newblock Direct language model alignment from online ai feedback.
\newblock {\em arXiv preprint arXiv:2402.04792}, 2024.

\bibitem{dubey2024llama}
Abhimanyu Dubey, Abhinav Jauhri, Abhinav Pandey, Abhishek Kadian, Ahmad Al-Dahle, Aiesha Letman, Akhil Mathur, Alan Schelten, Amy Yang, Angela Fan, et~al.
\newblock The llama 3 herd of models.
\newblock {\em arXiv preprint arXiv:2407.21783}, 2024.

\bibitem{kakade2002approximately}
Sham Kakade and John Langford.
\newblock Approximately optimal approximate reinforcement learning.
\newblock In {\em Proceedings of the Nineteenth International Conference on Machine Learning}, pages 267--274, 2002.

\bibitem{shridhar2020alfworld}
Mohit Shridhar, Xingdi Yuan, Marc-Alexandre C{\^o}t{\'e}, Yonatan Bisk, Adam Trischler, and Matthew Hausknecht.
\newblock Alfworld: Aligning text and embodied environments for interactive learning.
\newblock {\em arXiv preprint arXiv:2010.03768}, 2020.

\bibitem{wu2023autogen}
Qingyun Wu, Gagan Bansal, Jieyu Zhang, Yiran Wu, Shaokun Zhang, Erkang Zhu, Beibin Li, Li~Jiang, Xiaoyun Zhang, and Chi Wang.
\newblock Autogen: Enabling next-gen llm applications via multi-agent conversation framework.
\newblock {\em arXiv preprint arXiv:2308.08155}, 2023.

\bibitem{zhao2024expel}
Andrew Zhao, Daniel Huang, Quentin Xu, Matthieu Lin, Yong-Jin Liu, and Gao Huang.
\newblock Expel: Llm agents are experiential learners.
\newblock In {\em Proceedings of the AAAI Conference on Artificial Intelligence}, volume~38, pages 19632--19642, 2024.

\bibitem{sun2024adaplanner}
Haotian Sun, Yuchen Zhuang, Lingkai Kong, Bo~Dai, and Chao Zhang.
\newblock Adaplanner: Adaptive planning from feedback with language models.
\newblock {\em Advances in Neural Information Processing Systems}, 36, 2024.

\bibitem{krakovna2020specification}
Victoria Krakovna.
\newblock Specification gaming: the flip side of ai ingenuity.
\newblock DeepMind Blog, 2020.
\newblock Accessed: 2025-02-12.

\bibitem{weng2024rewardhacking}
Lilian Weng.
\newblock Reward hacking.
\newblock Blog post, 2024.
\newblock Accessed: 2025-02-12.

\bibitem{bradley1952rank}
Ralph~Allan Bradley and Milton~E Terry.
\newblock Rank analysis of incomplete block designs: I. the method of paired comparisons.
\newblock {\em Biometrika}, 39(3/4):324--345, 1952.

\bibitem{pan2022effects}
Alexander Pan, Kush Bhatia, and Jacob Steinhardt.
\newblock The effects of reward misspecification: Mapping and mitigating misaligned models.
\newblock {\em arXiv preprint arXiv:2201.03544}, 2022.

\bibitem{garg2021iq}
Divyansh Garg, Shuvam Chakraborty, Chris Cundy, Jiaming Song, and Stefano Ermon.
\newblock Iq-learn: Inverse soft-q learning for imitation.
\newblock {\em Advances in Neural Information Processing Systems}, 34:4028--4039, 2021.

\bibitem{xie2020q}
Tengyang Xie and Nan Jiang.
\newblock Q* approximation schemes for batch reinforcement learning: A theoretical comparison.
\newblock In {\em Conference on Uncertainty in Artificial Intelligence}, pages 550--559. PMLR, 2020.

\bibitem{choudhury2024better}
Sanjiban Choudhury and Paloma Sodhi.
\newblock Better than your teacher: Llm agents that learn from privileged ai feedback.
\newblock {\em arXiv preprint arXiv:2410.05434}, 2024.

\bibitem{swamy2023inverse}
Gokul Swamy, David Wu, Sanjiban Choudhury, Drew Bagnell, and Steven Wu.
\newblock Inverse reinforcement learning without reinforcement learning.
\newblock In {\em International Conference on Machine Learning}, pages 33299--33318. PMLR, 2023.

\bibitem{bagnell2003policy}
James Bagnell, Sham~M Kakade, Jeff Schneider, and Andrew Ng.
\newblock Policy search by dynamic programming.
\newblock {\em Advances in neural information processing systems}, 16, 2003.

\bibitem{osband2013more}
Ian Osband, Daniel Russo, and Benjamin Van~Roy.
\newblock (more) efficient reinforcement learning via posterior sampling.
\newblock {\em Advances in Neural Information Processing Systems}, 26, 2013.

\bibitem{swamy2022sequence}
Gokul Swamy, Sanjiban Choudhury, J~Bagnell, and Steven~Z Wu.
\newblock Sequence model imitation learning with unobserved contexts.
\newblock {\em Advances in Neural Information Processing Systems}, 35:17665--17676, 2022.

\bibitem{choudhury2018data}
Sanjiban Choudhury, Mohak Bhardwaj, Sankalp Arora, Ashish Kapoor, Gireeja Ranade, Sebastian Scherer, and Debadeepta Dey.
\newblock Data-driven planning via imitation learning.
\newblock {\em The International Journal of Robotics Research}, 37(13-14):1632--1672, 2018.

\bibitem{ross2011reduction}
St{\'e}phane Ross, Geoffrey Gordon, and J~Andrew Bagnell.
\newblock A reduction of imitation learning and structured prediction to no-regret online learning.
\newblock In {\em Artificial Intelligence and Statistics (AISTATS)}, 2011.

\bibitem{ross2014reinforcement}
Stephane Ross and J~Andrew Bagnell.
\newblock Reinforcement and imitation learning via interactive no-regret learning.
\newblock {\em arXiv preprint arXiv:1406.5979}, 2014.

\bibitem{sun2017deeply}
Wen Sun, Arun Venkatraman, Geoffrey~J Gordon, Byron Boots, and J~Andrew Bagnell.
\newblock Deeply aggrevated: Differentiable imitation learning for sequential prediction.
\newblock In {\em International Conference on Machine Learning (ICML)}, 2017.

\bibitem{abbeel2006application}
Pieter Abbeel, Adam Coates, Morgan Quigley, and Andrew Ng.
\newblock An application of reinforcement learning to aerobatic helicopter flight.
\newblock {\em Advances in neural information processing systems}, 19, 2006.

\bibitem{andrychowicz2020learning}
OpenAI:~Marcin Andrychowicz, Bowen Baker, Maciek Chociej, Rafal Jozefowicz, Bob McGrew, Jakub Pachocki, Arthur Petron, Matthias Plappert, Glenn Powell, Alex Ray, et~al.
\newblock Learning dexterous in-hand manipulation.
\newblock {\em The International Journal of Robotics Research}, 39(1):3--20, 2020.

\bibitem{agarwal2020model}
Alekh Agarwal, Sham Kakade, and Lin~F Yang.
\newblock Model-based reinforcement learning with a generative model is minimax optimal.
\newblock In {\em Conference on Learning Theory}, pages 67--83. PMLR, 2020.

\bibitem{liu2023agentbench}
Xiao Liu, Hao Yu, Hanchen Zhang, Yifan Xu, Xuanyu Lei, Hanyu Lai, Yu~Gu, Hangliang Ding, Kaiwen Men, Kejuan Yang, et~al.
\newblock Agentbench: Evaluating llms as agents.
\newblock {\em arXiv preprint arXiv:2308.03688}, 2023.

\bibitem{schick2023toolformer}
Timo Schick, Jane Dwivedi-Yu, Roberto Dess{\`\i}, Roberta Raileanu, Maria Lomeli, Luke Zettlemoyer, Nicola Cancedda, and Thomas Scialom.
\newblock Toolformer: Language models can teach themselves to use tools.
\newblock {\em arXiv preprint arXiv:2302.04761}, 2023.

\bibitem{zeng2023agenttuning}
Aohan Zeng, Mingdao Liu, Rui Lu, Bowen Wang, Xiao Liu, Yuxiao Dong, and Jie Tang.
\newblock Agenttuning: Enabling generalized agent abilities for llms.
\newblock {\em arXiv preprint arXiv:2310.12823}, 2023.

\bibitem{zhou2024archer}
Yifei Zhou, Andrea Zanette, Jiayi Pan, Sergey Levine, and Aviral Kumar.
\newblock Archer: Training language model agents via hierarchical multi-turn rl.
\newblock {\em arXiv preprint arXiv:2402.19446}, 2024.

\bibitem{cobbe2021training}
Karl Cobbe, Vineet Kosaraju, Mohammad Bavarian, Mark Chen, Heewoo Jun, Lukasz Kaiser, Matthias Plappert, Jerry Tworek, Jacob Hilton, Reiichiro Nakano, et~al.
\newblock Training verifiers to solve math word problems.
\newblock {\em arXiv preprint arXiv:2110.14168}, 2021.

\bibitem{luo2024improve}
Liangchen Luo, Yinxiao Liu, Rosanne Liu, Samrat Phatale, Harsh Lara, Yunxuan Li, Lei Shu, Yun Zhu, Lei Meng, Jiao Sun, et~al.
\newblock Improve mathematical reasoning in language models by automated process supervision.
\newblock {\em arXiv preprint arXiv:2406.06592}, 2024.

\bibitem{shao2024deepseekmath}
Zhihong Shao, Peiyi Wang, Qihao Zhu, Runxin Xu, Junxiao Song, Xiao Bi, Haowei Zhang, Mingchuan Zhang, YK~Li, Y~Wu, et~al.
\newblock Deepseekmath: Pushing the limits of mathematical reasoning in open language models.
\newblock {\em arXiv preprint arXiv:2402.03300}, 2024.

\bibitem{ma2023let}
Qianli Ma, Haotian Zhou, Tingkai Liu, Jianbo Yuan, Pengfei Liu, Yang You, and Hongxia Yang.
\newblock Let's reward step by step: Step-level reward model as the navigators for reasoning.
\newblock {\em arXiv preprint arXiv:2310.10080}, 2023.

\bibitem{wu2024inference}
Yangzhen Wu, Zhiqing Sun, Shanda Li, Sean Welleck, and Yiming Yang.
\newblock Inference scaling laws: An empirical analysis of compute-optimal inference for problem-solving with language models.
\newblock {\em arXiv preprint arXiv:2408.00724}, 2024.

\bibitem{havrilla2024teaching}
Alex Havrilla, Yuqing Du, Sharath~Chandra Raparthy, Christoforos Nalmpantis, Jane Dwivedi-Yu, Maksym Zhuravinskyi, Eric Hambro, Sainbayar Sukhbaatar, and Roberta Raileanu.
\newblock Teaching large language models to reason with reinforcement learning.
\newblock {\em arXiv preprint arXiv:2403.04642}, 2024.

\end{thebibliography}

% %%%%%%%%%% APPENDIX %%%%%%%%%%
% \newpage
% \appendix
% \part{Appendix} % Start the appendix part
% \input{appendix}

\end{document}